\documentclass[a4paper,fleqn]{cas-dc}

\usepackage[numbers]{natbib}
\usepackage{amsmath,amsfonts}
\usepackage{xcolor}
\usepackage{algorithmic}
\usepackage{algorithm}
\usepackage{array}
\usepackage{subfig}
\usepackage{float}
\usepackage{cuted}
\usepackage{capt-of}
\usepackage{textcomp}
\usepackage{stfloats}
\usepackage{url}
\usepackage{verbatim}
\usepackage{graphicx}
\usepackage{float}

\ExplSyntaxOn
\cs_gset:Npn \__first_foot:
{
  \parbox[t]{\textwidth}
  {
     \rule{\textwidth}{.2pt}\\
     \__first_footerline: \hfill Page~ \thepage {}
  }
}
\cs_gset:Npn \__cas_foot:
{
  \parbox[t]{\textwidth}
  {
   \rule{\textwidth}{.2pt}\\
   \sffamily\small
   \__first_footerline:
   \hfill Page~\thepage {}
  }
}
\ExplSyntaxOff

\def\tsc#1{\csdef{#1}{\textsc{\lowercase{#1}}\xspace}}
\tsc{WGM}
\tsc{QE}
\tsc{EP}
\tsc{PMS}
\tsc{BEC}
\tsc{DE}

\begin{document}
\let\WriteBookmarks\relax
\def\floatpagepagefraction{1}
\def\textpagefraction{.001}
\shorttitle{Clustering Node Attributed Networks with Graph Neural Networks and Self Learning}
\shortauthors{R.S. Luna and D.R Figueiredo }

\title [mode = title]{Clustering Node Attributed Networks with Graph Neural Networks and Self Learning}

\tnotetext[1]{This work received financial support through research grants provided by CNPq and FAPERJ (Brazil).}

\author[1]{Rodrigo de Sapienza Luna}
\cormark[1]
\fnmark[]
\ead{rluna@cos.ufrj.br}

\credit{Conceptualization of this study, Methodology, Software, Writing}

\affiliation[1]{organization={Systems Engineering and Computer Science (PESC), Federal University of Rio de Janeiro (UFRJ)},
                addressline={ Technology Center, Block H, Room 319},
                postcode={68511},
                city={Cidade Universitária},
                state={Rio de Janeiro},
                country={Brazil}}

\author[1]{Daniel Ratton Figueiredo}
\cormark[1]
\fnmark[]
\ead{daniel@cos.ufrj.br}
\ead[URL]{https://www.cos.ufrj.br/~daniel/}
\credit{Supervision, Conceptualization, Methodology, Writing}

\cortext[cor1]{Corresponding author}

\begin{abstract}
Graph clustering -- partitioning the node set of a graph into disjoint subsets that reflect some latent information -- is a fundamental problem as it finds applications in a myriad of different scenarios. While this classic problem has been tackled for decades by different communities, a recent variation of the problem driven by real data considers the scenario where nodes have attributes that are also informative. This has triggered novel methods that simultaneously leverage network information (edges) and node information (attributed) in the design of novel clustering algorithms. This work proposes a novel framework that builds on prior works that have applied graph neural networks (GNN) to graph clustering. The proposed framework operates in rounds of self learning in a fully unsupervised setting. In each round, a GNN generates representations for nodes that are used to cluster the nodes. This clustering influences the graph used to generate the node representation in the next round. Moreover, a context graph built in each round using the original graph is used to generate the node representations. Empirical results show that the proposed methodology extracts information from both network edges and node attributes in synthetic data, outperforming algorithms focused solely on the network or attributes when neither are very informative. Multiple rounds of learning also improve the performance and always outperforms a long single round of training (i.e., classic GNN graph clustering). When considering real datasets, empirical results indicate that the proposed methodology is competitive to state-of-the-art methods when cluster sizes are balanced.
\end{abstract}


\maketitle

\section{Introduction}
\label{sec:intro}
Graph clustering or graph partitioning or community detection consists of partitioning the set of nodes of a graph into disjoint subsets that reflect some higher order structure (e.g., many more edges within the subsets than across subsets)~\cite{schaeffer2007graph,fortunato2016community}. This is a fundamental problem in data science and network science since it finds applications in a myriad of contexts such as social networks, biological networks and information networks~\cite{newman2018networks}. Not surprising, graph clustering has been investigated for decades by various communities using different problem formulations and approaches (e.g., fast heuristics for minimizing modularity in network science, or approximation algorithms for optimal ratio-cut in graph theory).

A related but even more fundamental problem is data clustering where a set of points in some space must be partitioned into disjoint subsets that reflect some higher order structure (e.g. points in a subset are much closer than points in different subsets)~\cite{han2022data}. Data clustering finds even more applications as the input are just data points (and not a network) and an extensive literature in various communities. Indeed, the k-means algorithm was proposed over 50 years ago and is perhaps the most ubiquitous approach to tackle data clustering.

While both graph and data clustering are classic, only recently have they been unified into a single formulation~\cite{deshpande2018contextual,stanley2019stochastic}. In attributed networks, each network node is associated with some feature that is representative of the node (e.g., the DNA of the person in a social network). Suppose such feature is a point in some space. Note that nodes of this network can be clustered using just the node attributes (data clustering), just the network structure (graph clustering), or both kinds of information (joint data and graph clustering). This work considers the latter problem.

Jointly using node attributes and network structure to cluster nodes of an attributed network requires mixing evidence from the attributes with evidence from the network structure. Intuitively, such approaches should outperform any approach that uses just one kind of information, and performance gains should be higher when both kinds of information are relatively noisy (and independent). In this scenario, the joint approach can use more reliable attribute information to correct for noisy network information, and vice-versa. Of course, designing methodologies and building algorithms that can leverage this intuition is not trivial, and has been an active area of research over the past few years~\cite{deshpande2018contextual,stanley2019stochastic,zhang2021spectral,muller2023graph}.

An important consideration concerning graph or data clustering is prior information, or supervision. In a supervised setting, the labels of nodes or data points are available and indicate their clusters (or partially available in a semi-supervised setting). This information can be used to train a model that determines the cluster of unseen nodes. In contrast, in an unsupervised setting there is no prior information concerning the clusters. In modern and large problem instances, it is often the case that label information is not available (or is not reliable). This work considers the unsupervised scenario. The sole input to the problem here considered is a single instance of an attributed network.

Graph Neural Network (GNN) can be interpreted as a framework to generate representation for nodes using both the network structure and node attributes~\cite{liu2022introduction}. Such representations are points in some dimensional space and reflect some higher order structure of the network nodes (their local network structure and attributes). GNN has been successfully applied to node classification and link prediction in various scenarios. Not surprising, GNN has also recently been applied to clustering nodes of attributed network (see Section~\ref{sec:related}). In a nutshell, the representations generated by the GNN can be used as the input to a data clustering problem. However, when representations are generated as to optimize a generic objective function, such naive approaches often do not exhibit good performance. Most alternative solutions use a supervised (or semi-supervised) scenario where the label information can be used in the objective function or in some other stage. More recently, fully unsupervised approaches have also been proposed (more details in Section~\ref{sec:related}).

This work proposes a novel fully unsupervised framework for clustering attributed networks, called Dynamic Context Self-Learning Graph Neural Network (DCSL-GNN). The framework operates in rounds and its main contribution is the self-learning component and  generation of the context over the rounds. Using the representation and clusters of the previous round as well as the original network, the context (a set of nodes of fixed size) for each node is randomly generated. This context is used to generate a representation for nodes via a GNN (where "neighbors" are the nodes in the context). These representations are the input to a data clustering problem in Euclidean space, that can be solved with the classic k-means algorithm. These clusters and representations are then used to determine the context graph for the next round and their new representations. In a nutshell, representation and clusters found in the previous round influence the context generated for the nodes in the current round that is used by a GNN to generate new node representations, that are then used to determine the clusters of the current round.

The multiple rounds allows the framework to learn the right context for the nodes, leading to a better representation and consequently to a more effective clustering. In essence, DCSL-GNN employs a self-reinforcement mechanism: better representations generate better clusters, while better clusters generate better representations. The idea of node context is related to attention mechanisms in GNNs where edges incident to a node are weighted to focus on a subset of its neighbors. However, in DCSL-GNN any node can be in the context of another node, and not only a subset of its neighbors (as in attention mechanisms). This allows for nodes that are central in a cluster to be in the context of other nodes that belong to this cluster even when they are not neighbors in the original graph.

The framework is implemented and results on synthetic networks indicate its effectiveness in extracting information from both the network and attributes to generate effective clusters across the rounds of self-learning. Results clearly indicate that multiple rounds of self-learning are beneficial, and performance can be good even when one of the sources of information is very noisy (network or attributes). An evaluation using three benchmark datasets, DCSL-GNN outperforms all state-of-the-art algorithms in one of the datasets, indicating its potential in delivering good performance in the wild.

\section{Related Work}
\label{sec:related}

In this section, we review the related work in two categories that are related to our framework.

\subsection{Community Detection}

Community  detection (or graph clustering) is one of most important topics of modern network science, with practical application across various domains. Researchers from different fields have explored this problem for many years, employing various problem formulations and approaches, such as fast heuristics for minimizing modularity in network or approximating algorithms for optimal ratio-cut in graph theory~\cite{fortunato2010community}.

Modularity is a metric to measure the quality of a community (cluster) partition of a network using the difference between the number of edges within communities and the expected number of such edges across all pairs of nodes~\cite{newman2004finding}. Modularity optimization is the most used technique in community detection, and Newman and Girvan~\cite{newman2004finding} proposed the first heuristic to find communities that optimize modularity. Another popular and fast community detection algorithm is the Louvain algorithm proposed by Blondel et al.~\cite{Blondel_2008} which employs a greedy and efficient strategy to optimize the network modularity.

\subsection{Graph Embedding via Random Walks}

Graph embedding refers to the problem of generating a vector representation for nodes of a graph such that vectors capture some graph property of the nodes~\cite{cui2018survey,xu2021understanding}. This technique has been successfully applied to node classification and link prediction. Most of the techniques rely on generating a context for each network node often using short random walks that start at each node and considering a short sequence of visited nodes (window), such as DeepWalk, node2vec and struc2vec~\cite{cui2018survey,xu2021understanding}. This approach has been be applied to graph clustering by simply performing data clustering on the vector representation of the nodes. However, this approach does not directly consider node attributes and has been shown to have inferior performance~\cite{muller2023graph,shchur2019overlapping}.

While the framework here proposed relies on random walks over a weighted graph (where weights are determined from the previous round), the sequence of visited vertices are used to determine the relative importance of visited nodes and not to determine the context within a short sequence (window). More details in Section~\ref{sec:framework}.

\subsection{Graph Neural Network for clustering}

Recently, Graph Neural Networks (GNNs) have been introduced to address community detection on attributed graphs. One popular approach is employ a GNN in the graph reconstruction task (with node attributes) and use the embedding of the nodes in a classic data clustering task~\cite{bo2020structural,wang2022deep,shchur2019overlapping}. In this approach, the GNN serves as node encoders that leverages both structural and node attribute information. The learning process is achieved through the decoder process, where the model parameters are learned by comparing the difference between the original graph and the reconstructed graph (graph reconstruction with node attributes). After obtaining the node embedding through the encoder-decoder process, these embeddings are typically used as input to a data cluster algorithm to detect communities within the graph. For example, Wang et al.~\cite{wang2022deep} utilize a GNN encoder-decoder architecture for community detection. Bo et al. ~\cite{bo2020structural} developed a graph reconstruction-based clustering method that employs dual self-supervised learning to optimize the model jointly with the encoder-decoder. Shchur et al. apply a similar framework but that leverages the minimization of the likelihood function to detect overlapping communities generated by the GNN~\cite{shchur2019overlapping}.

Another popular unsupervised approaches adopt contrastive methods with specific loss functions to detect the network communities. The loss function reflects the quality of the communities using graph properties as well as a regularization term to reduce the number of communities.
Zhang et al.~\cite{zhang2021spectral} introduced SENet, which utilizes a spectral clustering loss to capture the global cluster structure and learn node embedding. CommDGI~\cite{zhang2020commdgi} proposes a combined objective of three loss functions: mutual information, modularity, and contrastive loss. DMoN~\cite{muller2023graph} is another framework that learns the model parameters through a loss function based on modularity but introduces a novel regularization term (linear in number of communities). Additionally, SEComm~\cite{bandyopadhyay2021unsupervised} combines the self-expressiveness with a self-supervised GNN for community detection through the optimization of two combined loss function. Last, DCOM-GNN~\cite{yang2023dcom} proposes a methodology to enhance the embeddings generated by an arbitrary GNN. The idea is to learn weights and matrices that manipulate the original GNN embedding to better discriminate inter-cluster and intra-cluster representations.

Our work introduces a novel approach to learning node representations for community detection in attributed graphs. Unlike most methods in the literature, the proposed framework does not rely on specific loss functions related to clustering such as modularity. Instead, we employ a dynamic graph and a self-learning module that iterative optimize the node embeddings through rounds. This approach allows to refine the dynamic graph based on the evolving node representations, enhancing community detection performance.

\section{Framework}

\begin{figure*}
\centering
\includegraphics[width=140mm]{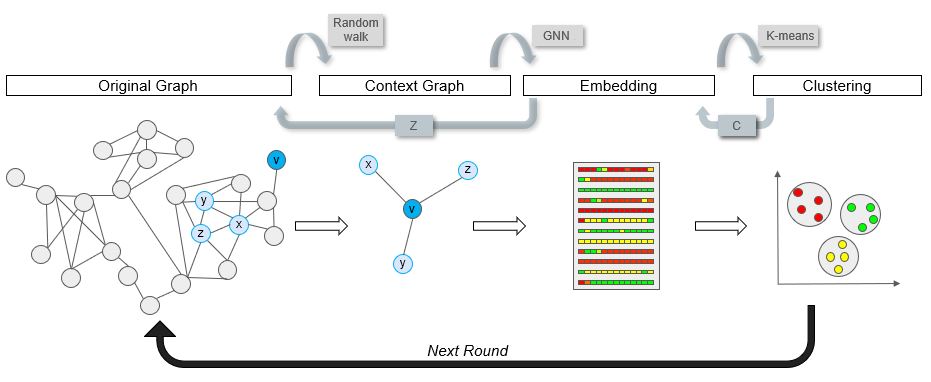}

\caption{Schematic representation of DCSL-GNN framework. Note that the clustering result is used to determine the edge weights of the original graph in the next round that is then used to generate the next context graph. Thus, embeddings influence clustering, and clustering influences embeddings.}
\label{fig:schema}
\end{figure*}

\label{sec:framework}
In this section, we present our proposed framework Dynamic Context Self-Learning Graph Neural Network (DCSL-GNN). As shown in Fig~\ref{fig:schema} our model contains three main modules:  Dynamic Context, Embedding and Clustering.

These modules are jointly learn in end-to-end manner to benefits each other.
\subsection{Problem Formulation}
Given an undirected attributed graph $G = (V,E,X)$, where $V = \{v_1,v_2,\dots,v_n \}$ is a set of n nodes and $E$ is a set of edges with $e_{ij} = (v_i,v_j) \in E$ if an edge exists between nodes $v_i$ and $v_j$, and $X = \{x_1, x_2, \dots, x_n\} \in \mathbb{R}^{n \times F}$ is the attribute matrix, where each node $v_i$ is associated by an attribute vector $x_i$ with dimension F.
Our goal is to partition the nodes into disjoint sets accordingly to the topological structure and the nodes attributes. Thus, we want to learn a function $f: V \rightarrow C$ where $C = \{C_1, C_2, \dots, C_K\}$ is the set of communities (or clusters), which map each node into a community $C_i$. We want to achieve the partitioning in the unsupervised way, i.e, without having any ground truth community information of a node, just knowing the number of communities.

Important notation is described in Table~\ref{tab:notacao}.

\begin{table}[ht]
\centering
\resizebox{0.5\textwidth}{!}{
\setlength{\tabcolsep}{8pt}
\fontsize{8}{4}
\begin{tabular}{|l|l|}
\hline
Notations & Definition \\ \hline
$G = (V,E,X)$         & Attributed graph \\ \hline
$G_c = (V_c,E_c,X)$       & Context graph  \\ \hline
$X \in \mathbb{R}^{n \times F}$         &  Node feature matrix\\ \hline
$H(\cdot) $      &   Embedding for node $v$  \\ \hline
$W $              &   Learning matrix \\ \hline
$ C^l \in \mathbb{R}^{K}$      & Community sets at $l$-th round \\ \hline
$||X,Y|| $      & Euclidean distance between $X$ and $Y$ \\ \hline
$L(\theta)$ & Triplet loss function \\ \hline
$K$         & Number of communities \\ \hline
$N(v)$      & Neighborhood of node $v$ \\ \hline
$s(v)$      & Silhouette value of node $v$ \\ \hline
$n_e$       & Number of epochs \\ \hline
$n_u$       & Mini-batch size \\ \hline
$n_w$       & number of random walks at each node \\ \hline
$n_s$       & number of random walk steps \\ \hline
\end{tabular}
}
\caption{Notation.}\label{tab:notacao}

\end{table}

\subsection{Dynamic Context}

GNNs aim to learn each node's embeddings by integrating their attributes with graph structure. In GNN, a layer can be viewed as a message passing between nodes, where each node updates its latent representation by aggregating the messages from its direct neighbors.

In the context of the graph clustering problem, nodes most representative of the cluster for a given node $v$ are not necessarily its neighbors. Suppose that the most central node $u$ for a cluster is $h$ hops away from a given node $v$. Then, the GNN should have at least $h$ layers for the information about $u$ to reach node $v$. Without this depth, the information propagation in the GNN is constrained and can hinder performance, resulting in a phenomenon known as under-reaching \cite{Barceló2020The}. However, unlike conventional neural network models, when additional layers are stacked in GNNs, the representation of the nodes becomes increasingly noisy. This leads to a significant degradation in prediction accuracy and overall performance, a problem known as over-smoothing \cite{kipf2016semi, balcilar2021analyzing,li2018deeper}. For that reason, most GNN models have very few layers, such as 2 or 3, and often deploy mechanism to avoid over-smoothing, such as dropout.

In order to avoid under-reaching and over-smoothing, our proposal modifies the  graph that over which the GNN will be executed. The original node neighborhood may not adequately represent the propagation of cluster signals. Thence, we design a flexible neighborhood sampling strategy biased by the similarity of nodes in the representation space. To capture cluster information, we add virtual edges to the closest nodes of cluster center of mass (in representation space). This approach allows both the information from cluster center through virtual edges and local structure of a node from original edges to be directly aggregated into a node, this new graph built is known as Context Graph, $G_c$.  As mentioned, DCSL-GNN is a self-learning framework trained in end-to-end manner across multiple rounds. This means in order to achieve better clustering performance, the nodes embedding and label assignments in previous round are employed to guide the optimization procedure of current round. In particular, the parameters obtained in the current round are use to generate the context of nodes in next round.

\begin{figure}[pos=h]
    \centering
    \begin{tabular}{@{}ccc@{}ccc@{}}
    \subfloat[Original Graph]{
      \includegraphics[width=0.40\linewidth]{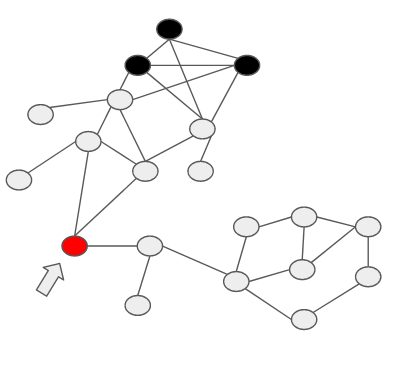}
    }
    \subfloat[Random walk perspective]{
      \includegraphics[width=0.50\linewidth]{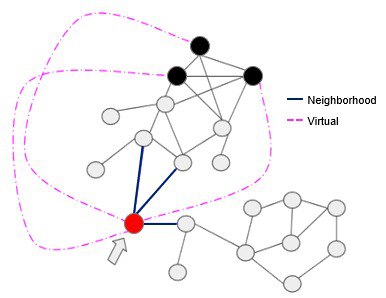}
    }
    \end{tabular}
    \caption{In each step of the random walk, a decision is made to either follow a virtual edge towards the cluster center nodes or explore the neighborhood of the current node. For each node, a total of $n_w$ short independent random walks are generated each taking $n_s$ steps.}
    \label{fig:virtual_edges}
\end{figure}

DCSL-GNN utilizes random walks as a mechanism to generate nodes that will then be used to determine the context of a given node. The original graph is augmented with edge weights that will bias the random walks. Each edge weight reflects the degree of similarity of the two nodes terms of their cluster membership (higher value means more similar). Moreover, virtual edges are added to the original graph, allowing the random walk to jump directly to nodes that are not neighbors of the current node. Figure~\ref{fig:virtual_edges} illustrates the idea of virtual edges, where at each step of the random walk, a decision is made to either traverse an edge in original graph (with probability $p$) or traverse a virtual edge (with probability $1-p$). Virtual edges takes the walker to a node that is close to the center of mass of the cluster. If a virtual edge is selected, the node to which the random walk transitions is chosen uniformly at random from the set of nodes closest to the cluster's mass center, with the $|CC(\cdot)|$ defined by a parameter. Otherwise, when selecting an original edge, the random walk explores the neighborhood based on proximity in the embedding space, resulting in a biased random walk where edge weights are inversely proportional to distances between the node representations.

To guide the random walk in the original graph, an edge weight function denoted by $w_H(\cdot)$ use the nodes representations generated by the previous round of DCSL-GNN, to bias the walk. It is worth noting that in the first round (no prior information available), transitions are made uniformly within the node's neighborhood. The $w_H(\cdot)$ value rises as the Euclidean distance between node representations diminishes. This characteristic steers the random walk towards nodes that are more inclined to connect to nodes with similar embedding, ensuring that the exploration of the random walk focuses on nodes that are closer in the embedding space.

The following equation determines the edge weight function:
\begin{equation}
    w_H(v,u) = \max_{z \in N(v)}(||H(v) - H(z)||) - ||H(v) - H(u)|| \; ,
    \label{peso}
\end{equation}
where $H(\cdot)$ is the node representation in euclidean space. Note that the value $\max_{z \in N(v)}||H(v) - H(z)||$ represents the largest distance from node $v$ to its neighbors. The value tends to be higher for nodes closer to $v$, approaching  $\max_{z \in N(v)}||H(v) - H(z)||$, and tends to be lower for nodes that are farther away, being zero for the farthest node. In essence, the weight function $w_H(\cdot)$ corresponds to the Euclidean distance relative to node $v$ and will bias the random walks.

Given a node $u$, a total of $n_w$ short  and independent random walks are simulated, each random walk taking exactly $n_s$ steps. In order to be clear, the transition probability of the random walks is given as follows. Let $c_i$ denote the $i$-th node visited by the walk, starting with $c_0 = u$. Nodes $c_i$ are generated by the following distribution:
\begin{equation}
\resizebox{.95\hsize}{!}{$
    P( c_i = k | c_{i-1} = v) =
    \begin{cases}
			w_H(v,k)/{Z_v}, & k \in N(v) \; \text{w.p. } p \\
            1/|CC(v)|     , & k \in CC(v) \; \text{w.p. }1-p
		 \end{cases}
   $}
    \label{eq:transicao}
\end{equation}
where $CC(u)$ is the set of nodes nearest to the cluster mass center of node u, and $Z_v$ is the normalization constant for vertex $v$ given by:
\begin{equation}
    Z_v = \sum_{i \in N(v)} w_H(v,i)
\end{equation}
Note that for each node $u$, the random walks starting from $u$ visit a total of $n_s n_l$ nodes (with repetition). Moreover, some nodes will be visited multiple times and each node visited by the walkers will have a visit count. The visit count will be used to determine the context graph, as follows.

The neighborhood of node $u$ in the Context Graph $G_c$, denoted as $N_c(u)$, is determined using the idea of importance-based neighborhoods, as originally proposed in the PinSage algorithm~\cite{ying2018graph}. This method defines the neighborhood of a node $u$ as the the nodes that exert the most influence on node $u$. Thus, the most visited nodes by the random walks starting at $u$ will be the neighbors of node $u$. Since each node $u$ is associated with an independent non-identically visit count distribution, we employ a threshold value to determine if a visited node will be included in its neighborhood. This threshold value is defined as the $k$-th percentile of the visit distribution. Consequently, $N_c(u)$ is comprised of nodes that have been visited at least the value of the $k$-th percentile. Note that this will generate nodes with different degrees in the context graph, since the visit count distribution can be quite different among different nodes.

Note that constructing node neighborhoods based on importance has several advantages. First, the selection of the most important nodes is based by visits counts, it enables nodes with high visit counts being direct neighbors in the $G_c$, even when the node is many hops away in the original graph. This will reduce the noise and amplify the signal of important nodes. Second, a few layers stacked in GNN in the original graph could already incur unwelcome messages between clusters, however this is less likely to occurs in the context graph, because the neighborhood contains only similar nodes that are likely to belongs to the same clusters. Third, as only the most important nodes are chosen, it allows to reduce the neighbourhood size without loss of generality and helps to decrease the algorithm's memory consumption.

The framework can handle graphs that are not connected, as the neighborhood for a node in $G_c$ will be generated independently for all nodes. Thus, multiple connected components in the original graph will be represented by multiple connected components in the context graph. However, the context graph can also have more connected components than the original graph. Nevertheless, nodes in different connected components in the $G_c$ can be part of the same cluster, as the node attributes play an important role when generating node representations.

\subsection{Weights in Context Graph}

Edges of the context graph will also have weights that will better guide the learning of the representation by the GNN. In the first round, all edges in $G_c$ are assigned a weight value of $1$, implying that all messages from the neighborhood hold equal importance during the aggregation step. However, due the stochastic nature of the context graph, uncertain edges may be included in the neighborhood, potentially introducing noise and mixing their latent representations. Furthermore, nodes that exhibit high similarity in their attributes should contribute with greater importance during the aggregation step of the GNN. Thus, an edge $(u,v)$ in $G_c$ is deemed uncertain when the attributes from nodes $u$ and $v$ lack similarity and at least one of the nodes has an attribute that is highly similar to the attributes of the cluster to which it belongs.

The silhouette coefficient algorithm~\cite{rousseeuw1987silhouettes} is employed to quantify the cohesion of attributes within a cluster. This coefficient measures the similarity of a data point (node attribute) to its own cluster compared to other clusters. It considers both the distance between the data point and other points within its own cluster (cohesion) and the distance between the data point and points in other clusters (separation).

Assuming the node embeddings have been clustered into $C$ clusters using some clustering algorithm, the silhouette coefficient for each data point $i$ can be defined as follows:
\begin{equation}
    s(i) = \frac{b(i) - a(i) }{ \max(a(i),b(i))} \; ,
\end{equation}
where $a(i)$ is the average intra-cluster distance, i.e., the average distance between each point within the cluster of node $i$, and $b(i)$ is the average inter-cluster distance, which is the average distance between the node $i$ and all points in other clusters, defined as follows:
\begin{equation}
    a(i) = \frac{1}{|C_{z_i}| - 1  } \sum_{j\in C_{z_i}, i \neq j} d(i,j)
\end{equation}
\begin{equation}
    b(i) = \min_{y \neq z_i} \frac{1}{C_y} \sum_{j \in C_y} d(i,j)
\end{equation}
where $C_{z_i}$ is the cluster of node $i$ (set of nodes) and $z_i$ is the cluster number of node $i$.
\newline

For each node $u$ in a cluster, $-1\leq s(u)\leq 1$. A silhouette value closer to 1 indicates that the data is appropriately clustered, a value closer by -1 indicates poor clustering, and a value closer to 0 suggests the value is on the border between multiples clusters.

The dynamic nature of context graph enables the utilization of the previous clustering results to guide the optmization of the current round. DCSL-GNN determines the edge weights of $G_c$ in current process using the clusters obtained in the previous round. However, this optmization process might take a considerable amount of rounds to discover consistent clusters. Thus, an attribute $X[u]$ of node $u$ is considered properly clustered only if $s(u) \geq \beta$, where is $\beta$ is a threshold value. If one of the nodes from an edge satisfies this condition, the edge weight value will depend on the inverse Euclidean distance between the attribute values of the nodes, otherwise the weight is considered 1. The following equation determines the edge weight:
\begin{equation}
\resizebox{.85\hsize}{!}{$
    w(v,u) =
    \begin{cases}
        \min\left( \frac{1}{\lVert X[u], X[v]\rVert }, \alpha \right), & \text{if } s(u) \geq \beta \text{ or } s(v) \geq \beta \\
        1
    \end{cases}
    $}
\end{equation}

An edge weight $w(v,u)$ with nodes attributes $X[v]$ and $X[u]$ assumes values between $[0,\alpha]$. It is important to note that $\alpha$ is a parameter which controls the upper bound of the weight, in order to prevent the edge weight from approaching infinity and dominating the aggregation in the GNN. Therefore, when $w(i,j) < 1$, the edge is penalized as it reduces the importance between the nodes. Conversely, when $w(i,j) > 1$, it increases the influence in the messages between them. Finally, when $w(i,j) = 1$ the  clustering of the nodes is not reliable and the edge is neutral. Last, these edge weights is part of the input to the GNN, along with $G_c$ and the node attributes.

\subsection{Embedding}

The encoder proposed in~\cite{kipf2016semi} was adopted in the embedding module of the framework. Given the context graph $G_c$ with a adjacency matrix $A$ a stack layer of the model is defined as follow:
\begin{equation}
    H^{\left(l+1\right)} =  \sigma \left( \Tilde{D}^{-\frac{1}{2}} \Tilde{A} \Tilde{D}^{-\frac{1}{2}}  H^{\left(l\right)} W^{\left(l\right)}  \right)
    \label{eq:GNN_1}
\end{equation}
where $\Tilde{A} = A + I$, where $I \in \mathbb{R}^{n \times n}$ is the identity matrix with goal to add self-loops and the degree diagonal matrix with $\Tilde{D}_{ii} = \sum_{j \in V} \Tilde{A}_{ij}$. The matrix $W^{l}$ of the $l$-th layer is a trainable parameter in GCN and $\sigma$ is the activation function ($ReLu$ in the experiments). The last layer of embedding matrix $H^l \in \mathbb{R}^{n \times F}$ is the representation matrix and contains in each row the corresponding node representation with dimension $F$.

The operator can be defined in the perspective of a node as follow:
\begin{equation}
   H_i^{\left(l+1\right)} = W^{\left(l\right)} \sum_{j\in N(i)} \frac{w(j,i)}{\sqrt{ \Tilde{d}_j \Tilde{d}_i }} H^{\left(l\right)}_j
\end{equation}
where $w(j,i)$ denotes the edge weight from source node $j$ to target node $i$ and $\Tilde{d_j}$ denotes the node $j$ degree.

Note that $G_c$ is the input (along with node attributes) to learning the node representations in the
GNN. However, each round the framework builds a new context graph $G_c$. Thus, for each round, the learning of the node embedding uses on a potentially different graph. However, the parameters of the GNN model (i.e.,  the weight matrices $W^{\left(l\right)}$) are carried over from the previous round (and not re-initalized), and this reduces the training time of the GNN model in each round.

\subsection{Clustering}

We apply the K-means clustering algorithm on the data points provided by the Embedding module. Given a set observations $X = \{x_1,\dots,x_n\}$ this algorithm aims to partition the data points into $K$ disjoint sets $C= \{C_1,\dots,C_k\}$ with go to minimize the objective function bellow:

\begin{equation}
    L = \sum_{n=1}^{N} \sum_{k=1}^{K} r_{nk} ||x_n - \mu_k||^2
    \label{kmeans}
\end{equation}

K-means aims to optimize the parameters $\mu_k$ and $r_{nk}$ that minimizes the objective function.These parameters are, respectively, the mean data points that belongs to the cluster k and a variable indicator, as described, respectively, in Equations \ref{eq:mu} and \ref{eq:rnk}. Note that if one of the parameters is known, another one can be inferred. For this, an iterative method is applied in order to optimize one by one.

\begin{equation}
    \mu_k = \frac{\sum_n r_{nk} x_n}{\sum_n r_{nk}} , \text{for } k = 1,\ldots,K
    \label{eq:mu}
\end{equation}

\begin{equation}
    r_{nk} = \begin{cases}
      1, & \text{if } k = \arg\min_j \lVert x_n - \mu_j \rVert^2,\quad k = 1,\ldots,K,\\
      0, & \text{otherwise.}
     \end{cases}
     \label{eq:rnk}
\end{equation}

Standard optimization methods are used to solve the $K$-means problem in order to determine the cluster assignment matrix $r_{nk}$.

\subsection{Model training}

An important contribution of this work is the self-learning component in the proposed framework. In the training stage the process involves passing through each module,  Dynamic Context, Embedding, and Clustering, in iterative rounds. In each round, the model in the Embedding module is trained for a fixed number of epochs to optimize and learn the weights matrices and bias for each layer in GNN. For each epoch, (Context Graph) is considered, as opposed to a sample the entire network. However, we evaluate the loss function and update the parameters by $\frac{n}{n_b}$ times using mini-batches of size $n_b$ which are uniformly sampled without replacement from the graph nodes. Despite using all nodes and edges in message passing, only the nodes in the batch are used to evaluate the loss function at each mini-batch. This approach allows us to make efficient use of the entire network while still updating the model parameters in a batch-wise manner, an approach known to be lead to better convergence.
\par Algorithm~\ref{alg:alg1} illustrates the DCSL-GNN framework. In the very first round, before the clusters are formed, each node is considered to be in its own individual cluster (cluster $C^0$). Thus, during the first random walk procedure, each node in the neighborhood has an equal probability of being visited, and the walk can return to the given node (as the node is the cluster center) with probability $(1-p)$.

\begin{algorithm}[H]
\caption{DCSL-GCN: model training using mini-batches for the evaluation function}\label{alg:alg1}
\begin{algorithmic}
   \FOR{$r = 1, \ldots, n_r$}
   \STATE $ G_c \gets \textsc{GenerateContext($G,H^{r-1}, C^{r-1})$}$
   \FOR{$e = 1, \ldots, n_e$}
   \STATE $H^r = GenerateEmbeddings(G_c,W)$
   \FOR{$i = 1, \ldots, \frac{n}{n_b}$}
        \STATE $\textsc{S} \gets  \textsc{GenerateSeeds($V_c,n_u$)}$
        \STATE $\textsc{L} \gets  \textsc{ComputeLoss($S, H^r$)}$
        \STATE $\textsc{UpdateWeights(W,L)}$
     \ENDFOR
     \ENDFOR
     \STATE $C^r \gets \textsc{GenerateClusters($H^r$)}$
     \ENDFOR

\end{algorithmic}
\label{alg1}
\end{algorithm}

For training the GNN model, a loss function based on triplet loss was adopted, as shown in Equation~\ref{loss}. In this function a sample data point (known as anchor) is compared with the matching input (positive sample) and non-matching input (negative sample).
Nodes that belong to the same cluster as the anchor node are considered positive samples, denoted as $x_n^+$. Nodes that do not belong to the same cluster are considered negative samples, denoted as $x_n^-$. This function aims to minimize the distance between an anchor and a positive sample; and maximizes the distance between the anchor and a negative sample. The hyper-parameter $\Delta$ enforces a margin distance between positive and negative pairs.
\begin{equation}
    L\left(\Theta \right) = \frac{1}{N} \sum_{n=1}^{N} \max\left( ||x_n - x_n^+||^2 - ||x_n - x_n^-||^2 +\Delta      ,0\right)
    \label{loss}
\end{equation}

This loss function was adopted because it aims to minimize the distances between similar data points (positive pairs) and increase the distances between dissimilar data points (negative pairs).  This behavior is aligned with the procedure to construct the Context Graph, as the random walks are biased by the Euclidean distance in the latent space. Therefore, in every round, similar nodes in embedding space becomes closer and distinct nodes becomes farther apart away.  This process contributes to create a better context in the subsequent round.

Finally, the relatively simple GCN architecture proposed by~\cite{kipf2016semi} was adopted. It a simple model with only a few trainable parameters (only the weights and bias matrices). A GCN model with only two layers is considered, and thus, we need to learn two matrices, $W^{\left(0\right)}$ and $W^{\left(1\right)}$ as seen in the Equation~\ref{eq:GNN_1}. In each round of training, the parameters of the GCN for round $r$ are initialized with the parameter values obtained from the previous round $r-1$.
The computational cost of a simple model is reduced and the framework becomes relevant in scenarios where the graph has a large number of vertices and edges. Moreover, this procedure must be efficient since a model will be trained in every round of the proposed framework.

\section{Evaluation}
\label{sec:eval}

This section provides an evaluation of DCSL-GNN in various scenarios using synthetic attributed networks and benchmark datasets, as well as a direct comparison with other algorithms.

\subsection{Attributed Stochastic Block Model}

The Stochastic Block Model (SBM)~\cite{holland1983stochastic} is a generative model for random graphs that produces networks containing cluster structure. The model is defined by the number of clusters $K$, the number of nodes in each cluster, $n_i, i = 1, \ldots, K$, and a symmetric matrix $B \in [0,1]^{K \times K}$ specifying the connectivity probabilities between clusters. Thus, $B[i,j]$ is the probability that an edge is present between a pair of nodes $u$ and $v$ that belong to clusters $i$ and $j$, respectively. To model homophily and induce clusters, $B[i,i] > B[i,j], i \neq j$.

The classic SBM has been recently extended to incorporate a model for node attributes that can be correlated with network structure~\cite{stanley2019stochastic,deshpande2018contextual}. A simple model for node attributes assumes that all nodes in the same cluster have a random attribute drawn from the same distribution. In particular, the attributes of such nodes are independently and identically distributed. Gaussian distributions with parameters $\mu_i$ and $\sigma_i^2$, $i = 1, \ldots, K$, are used to model the attributes of nodes in cluster $i$. Note that, conditioned on its cluster, the edges incident to a node and its attribute are independent.

This model allows us to explore the role of network information, represented by edges, and node information, represented by attributes, in the identification of clusters. In particular, by increasing the variance $\sigma_i^2$ for all $i$, the node information becomes noisier and less reliable. Similarly, by decreasing $B[i,i]$ for all $i$, the network information becomes noisier and less reliable.

The Normalized Mutual Information (NMI) is a classic metric used to evaluate the quality of a clustering with respect to a ground truth. NMI values range from $[0,1]$, where values close to $0$ indicate poor cluster quality, while values approaching $1$ indicate high cluster quality. This metric is used to assess the performance of DCSL-GNN and the other algorithms.

To ensure statistical significance given the stochastic nature of DCSL-GNN and the synthetic networks, multiple independent experiments using different sampled networks were conducted for each scenario. The reported results are the mean values obtained from multiple runs, with 50 runs per scenario.

For the experiments, the generated networks have $K=4$ clusters with $n_i = 100$, $i=1,\ldots,4$, resulting in a total of $N=400$ nodes. The intra-cluster probability $B[i,i] = p_I$ is the same for all clusters and is varied in the experiments, while the inter-cluster probability is fixed at $B[i,j] = 0.1$, $i \neq j$. Thus, the average number of edges between a node and nodes in different clusters is 30, corresponding to 10 edges per cluster. The attributes for each cluster follow Gaussian distributions with means $[10,20,30,40]$, respectively, and have identical variance $\sigma^2$, which is varied in the experiments.

\begin{figure*}
\centering
\includegraphics[width=0.95\linewidth]{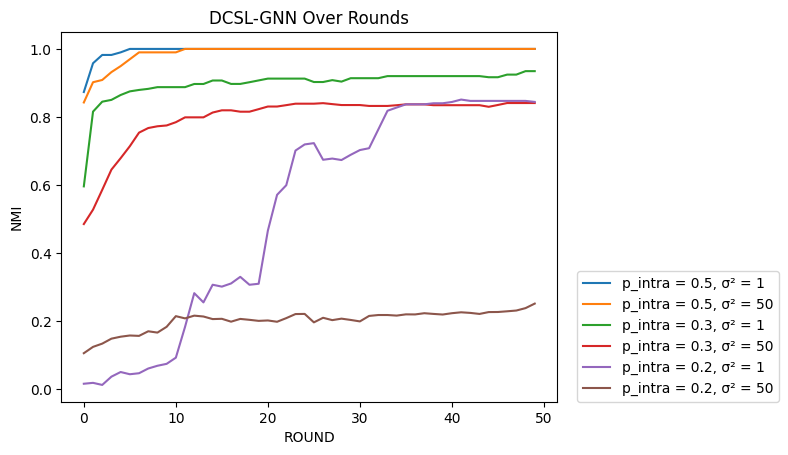}
\caption{Normalized Mutual Information over 50 rounds in different scenarios.}
\label{fig:learncapability}
\end{figure*}

\subsection{Self-learning potential}

DCSL-GNN utilizes a self-supervised mechanism to learn through multiple rounds. The goal is to understand the behavior of the model across multiple rounds under different scenarios of noise in the network structure and node attributes. Figure~\ref{fig:learncapability} shows the evolution of the NMI as a function of the learning rounds for the different scenarios. Note that in scenarios where the network information has little noise, namely $p_I = 0.5$, DCSL-GNN exhibits rapid convergence towards a perfect clustering after a few rounds of self-learning. This occurs even when the node information is very noisy, with $\sigma^2 = 50$. However, when the network information becomes noisier, with $p_I = 0.3$, the model's performance improves over the rounds of self-learning and converges, but not to a perfect clustering. This occurs even when node attributes have little noise, with $\sigma^2 = 1$. Finally, in the challenging scenarios where the network information is very noisy, with $p_I = 0.2$, and attributes have little noise, with $\sigma^2 = 1$, DCSL-GNN improves its performance over many rounds of self-learning, eventually converging and delivering good performance. However, when attributes are also very noisy, with $\sigma^2 = 50$, the multiple rounds of self-learning do not improve performance, and the model converges to low cluster quality, similar to a random guess. Indeed, in such a scenario, there is not enough information to perform effective clustering.

\begin{figure*}
\centering

\begin{minipage}{0.31\textwidth}
\centering
\includegraphics[width=\linewidth]{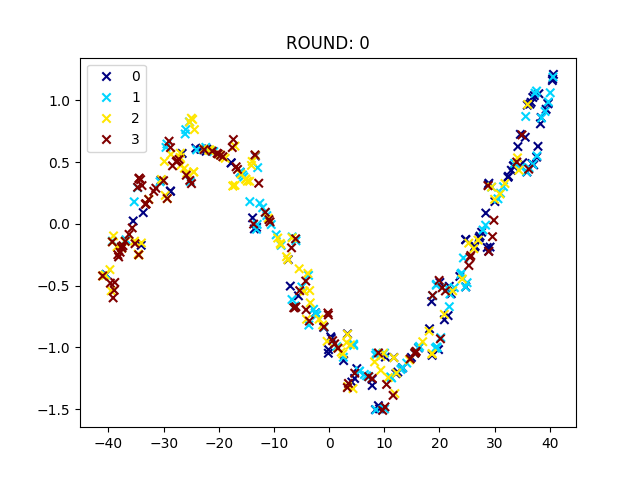}
\end{minipage}
\hfill
\begin{minipage}{0.31\textwidth}
\centering
\includegraphics[width=\linewidth]{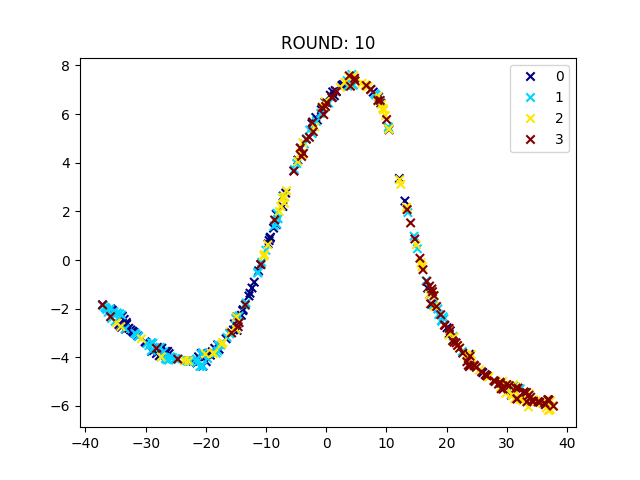}
\end{minipage}
\hfill
\begin{minipage}{0.31\textwidth}
\centering
\includegraphics[width=\linewidth]{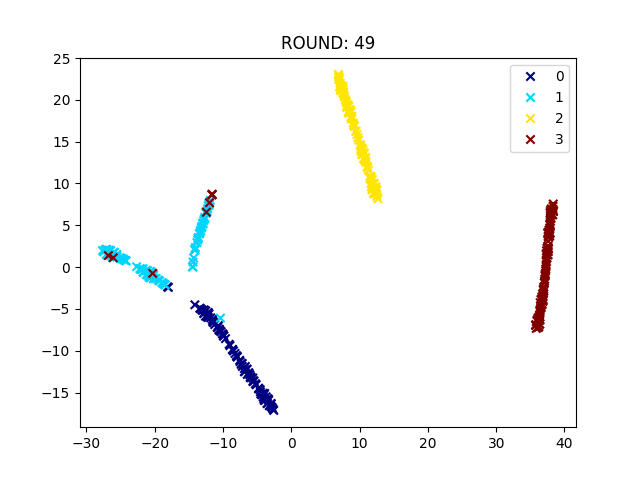}
\end{minipage}

\captionof{figure}{Projection of node embeddings at the end of different rounds for the scenario where $p_I = 0.2$ and $\sigma^2 = 1$ (two-dimensional projection via t-SNE).}
\label{fig:embeddings}
\end{figure*}

Figure~\ref{fig:embeddings} shows the t-SNE projection in two dimensions of the node embeddings generated by DCSL-GNN at the end of different rounds in the scenario where $p_I = 0.2$ and $\sigma^2 = 1$. Initially, the projections do not reflect any cluster structure. However, as the rounds progress, the clusters begin to form.

\subsection{Network and node noise}

The performance of DCSL-GNN in different scenarios of noise for the network and node attributes is evaluated. This is compared with the performance of algorithms that consider only the network structure, namely Louvain modularity, and only the node attributes, namely K-means, to cluster the data in the exact same networks. Additionally, DCSL-GNN is evaluated without the self-learning component, denoted as NSL in the results, which is trained with only one round but with the same total number of epochs as the self-learning model.

\begin{strip}
\centering

\begin{minipage}{0.32\textwidth}
\centering
\includegraphics[width=\linewidth]{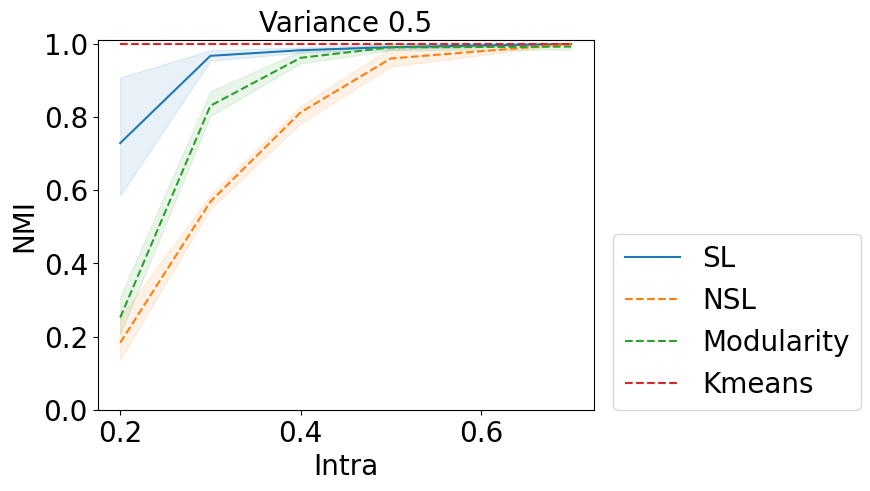}
\end{minipage}
\hfill
\begin{minipage}{0.32\textwidth}
\centering
\includegraphics[width=\linewidth]{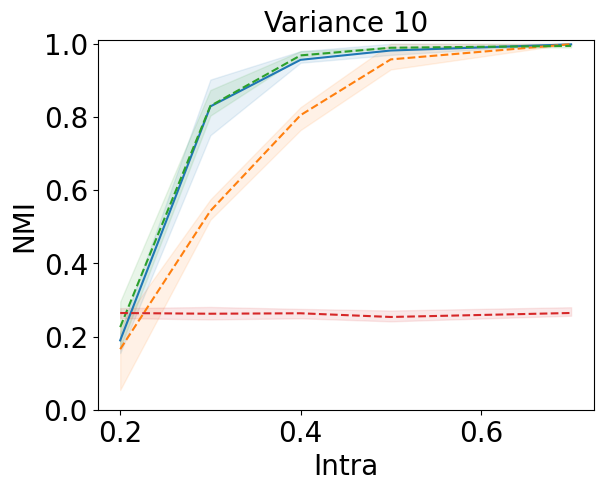}
\end{minipage}
\hfill
\begin{minipage}{0.32\textwidth}
\centering
\includegraphics[width=\linewidth]{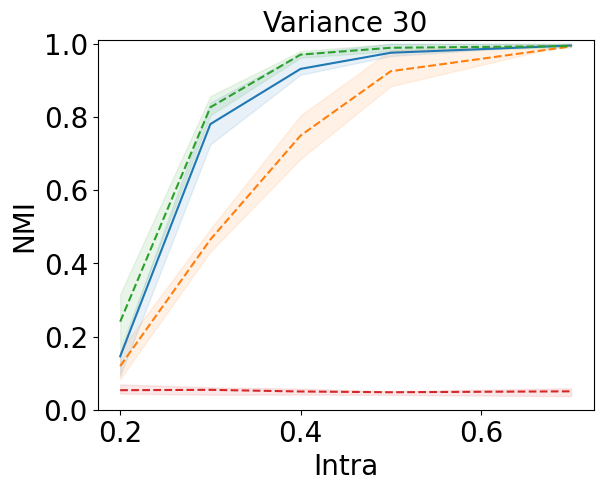}
\end{minipage}

\captionof{figure}{Performance, with mean and standard deviation shown as shaded regions, of different clustering algorithms under different scenarios as a function of $p_I$ for different attribute variances.}
\label{fig:varATT}
\end{strip}

Figure~\ref{fig:varATT} compares the performance of the algorithms under a fixed variance in each plot but varying intra-cluster probabilities. Observe that in regimes with high intra-cluster probability, namely $p_I \geq 0.4$, the model achieves similar results to the Louvain algorithm, even when the attributes are very noisy, with $\sigma^2 = 30$. Note that in all scenarios, the performance decays as the network information becomes noisier, that is, as $p_I$ is reduced. However, this decay depends on the quality of the attribute information, with better performance when the attribute information is less noisy, as in the case $\sigma^2 = 0.5$. Interestingly, DCSL-GNN consistently outperforms the model without the self-learning component in all scenarios, indicating the effectiveness of the self-learning mechanism in improving performance.

\begin{figure*}
\centering

\begin{minipage}{0.32\textwidth}
\centering
\includegraphics[width=\linewidth]{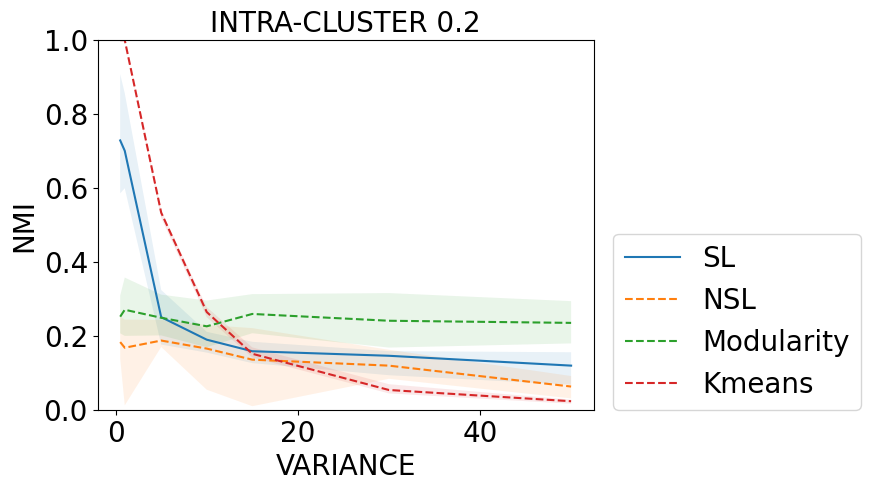}
\end{minipage}
\hfill
\begin{minipage}{0.32\textwidth}
\centering
\includegraphics[width=\linewidth]{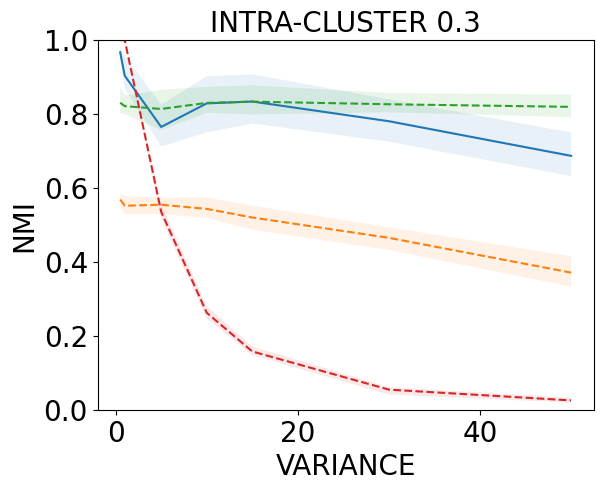}
\end{minipage}
\hfill
\begin{minipage}{0.32\textwidth}
\centering
\includegraphics[width=\linewidth]{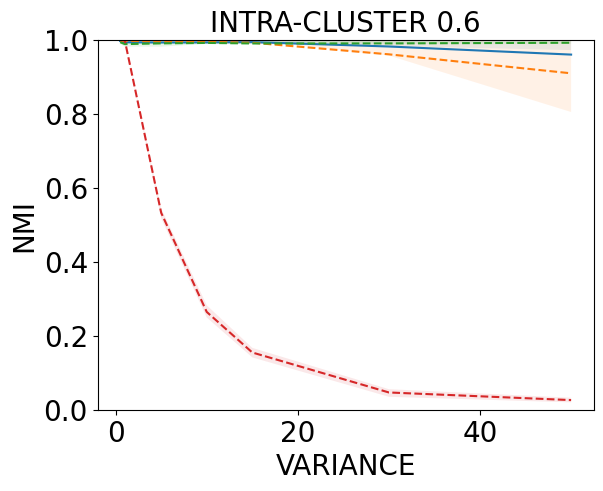}
\end{minipage}

\captionof{figure}{Performance, with mean and standard deviation shown as shaded regions, of different clustering algorithms under different scenarios as a function of attribute variance for different values of $p_I$.}
\label{fig:intraFixed}
\end{figure*}

Figure~\ref{fig:intraFixed} shows the result for a fixed intra-cluster probability while the attribute variance is varied. Note that in all scenarios, the performance of DCSL-GNN decays with the increase in the variance. However, this decay depends on $p_I$. When the network information has little noise, with $p_I = 0.6$, DCSL-GNN provides perfect clustering even under very high attribute noise, with $\sigma^2 = 50$. Note that DCSL-GNN outperforms K-means when attribute information is noisy but network information has little noise. Moreover, DCSL-GNN outperforms Louvain modularity when network information is noisy but attribute information has little noise. This indicates that DCSL-GNN is flexible and capable of extracting information from both the network structure and node attributes to perform more effective clusterings.

\subsection{Benchmark datasets}

Experiments were performed on three real citation networks that are commonly used to evaluate different network analysis algorithms: Cora, Citeseer, and PubMed. These datasets are widely employed in community detection studies, serving as benchmark graphs for evaluating the performance of various frameworks and methods. DCSL-GNN was executed 20 times with the hyperparameters manually tuned according to the dataset. Table~\ref{table:datasets1} describes important characteristics of these datasets.

\begin{table}[H]
\centering
\resizebox{0.95\linewidth}{!}{
\begin{tabular}{lllllllll}
\hline
Dataset & & Nodes & Edges & $F$ & $K$ & cc & $<k>$ & Imbalance \\
\hline
Cora     & & $2708$  & $5278$  & $1433$ & $7$ & $0.24$ & $3.89$ & $4.54$ \\
Citeseer & & $4230$  & $4732$  & $5337$ & $6$ & $0.11$ & $2.52$ & $1.48$ \\
PubMed   & & $19717$ & $44324$ & $500$  & $3$ & $0.06$ & $4.49$ & $1.91$ \\
\hline
\end{tabular}
}
\caption{Network characteristics of the benchmark datasets. $F$ represents the dimension of the node attributes, $K$ is the number of clusters, $cc$ denotes the average clustering coefficient of the network, $<k>$ is the average node degree, and imbalance is the ratio between the largest and smallest cluster.}
\label{table:datasets1}
\end{table}

Table~\ref{table:results} presents the evaluation of multiple frameworks for clustering attributed graphs, with the NMI metric used to measure cluster quality. Note that for each dataset, the performance of the different algorithms is very different. Moreover, the performance of each algorithm across datasets is also very different. In the Cora dataset, the MGCCN framework~\cite{liu2022multilayer} achieves the highest NMI value and is considered state-of-the-art. Similarly, in the PubMed dataset, SEComm~\cite{bandyopadhyay2021unsupervised} is considered state-of-the-art. Furthermore, the proposed framework DCSL-GNN shows superior performance for the Citeseer dataset. Clearly, there is no single algorithm that is best across all datasets.

\begin{table}[H]
\centering
\resizebox{0.95\linewidth}{!}{
\begin{tabular}{lllll}
\hline
 & & Cora & Citeseer & PubMed \\
\hline
K-means$^*$ & & $12.8 \pm 3.5$ & $26.54 \pm 2.3$ & $26.77 \pm 0.2$ \\
Louvain$^*$ & & $41.49$ & $28.62$ & $21.81$ \\
GAE~\cite{bandyopadhyay2021unsupervised} & & $40.69$ & $18.34$ & $22.97$ \\
GIC~\cite{mavromatis2021graph} & & $53.70$ & $45.30$ & $31.90$ \\
AGC~\cite{zhang2019attributed} & & $53.68$ & $41.13$ & $31.59$ \\
MGCCN~\cite{liu2022multilayer} & & \textcolor{red}{$60.20$} & $43.20$ & $-$ \\
DNENC-Att~\cite{wang2022deep} & & $52.80$ & $39.70$ & $26.60$ \\
DMoN~\cite{muller2023graph} & & $48.80$ & $33.70$ & $29.80$ \\
CommDGI~\cite{zhang2020commdgi} & & $57.90$ & $41.90$ & $35.70$ \\
SEComm~\cite{bandyopadhyay2021unsupervised} & & $56.04$ & $42.53$ & \textcolor{red}{$36.50$} \\
\hline
\textbf{DCSL-GNN}$^*$ & & $46.95 \pm 3.1$ & \textcolor{red}{$47.74 \pm 2.8$} & $24.27 \pm 1.20$ \\
\hline
\end{tabular}
}
\caption{Normalized Mutual Information (NMI) of clustering performance of different algorithms for Cora, Citeseer, and PubMed datasets. Performance values were obtained from the corresponding papers and were not reproduced in this work. The performance of algorithms marked by $^*$ was obtained in this work. Highlighted in red is the best result for each dataset, illustrating the lack of a single best algorithm across all datasets.}
\label{table:results}
\end{table}

While DCSL-GNN shows superior performance on the Citeseer dataset, its performance on the other two datasets is relatively far from the state-of-the-art. A possible cause for this lower performance is cluster, or class, imbalance, which can significantly impact the performance of DCSL-GNN. In particular, clusters with fewer nodes contribute less to the objective function, which can result in a bias towards the dominant cluster. Indeed, for the Cora dataset, DCSL-GNN shows very good accuracy for the dominant cluster but negligible accuracy for the smallest cluster, although these results are not shown here.
\section{Conclusion}
\label{sec:conclusion}

This paper proposed a novel fully unsupervised framework for clustering network nodes with attributes, called DCSL-GNN. The framework is based on two key ingredients: context graph and self-learning. The context graph is a weighted graph built from the original graph where edges indicate that nodes should belong to the same cluster (independent of their graph distance in the original graph). The self-learning allows DCSL-GNN to generate clusters and that generate new and better context graphs that can then be used to generate better clusters. A GNN is responsible for generating the node representations from the context graph which are in turn used by a clustering algorithm in Euclidean space. The GNN leverages the network and attribute information as well as a loss function that depends on clustering assignments of the previous self-learning round in order to generate meaningful representations.
This iterative self-learning approach enables DCSL-GNN to refine both the node representations and the clustering assignments in a mutually beneficial manner.

Using a network model with clusters and attributes, DCSL-GNN is evaluated and compared to other algorithms with results indicating its potential in extracting information from both the network and attributes to generate effective clusters. In particular, multiple rounds of self-learning are always advantageous, and network or attribute information can be leveraged even when the other is very noisy. Finally, when considering benchmark datasets, DCSL-GNN achieved mixed results being superior to all state-of-the-art algorithms for one dataset.

\printcredits

\section*{Declaration of competing interest}
The authors declare the following financial interests/personal relationships which may be considered as potential competing interests: Daniel Ratton Figueiredo reports financial support was provided by National Council for Scientific and Technological Development (CNPq), Brazil.

\section*{Data availability}
Data will be made available on request.
\bibliographystyle{cas-model2-names}

\bibliography{refs}

\bio{}\textbf{Rodrigo de Sapienza Luna}
received a BS in Computer Science from the Federal University of Rio de Janeiro (UFRJ), Brazil, in 2020. Subsequently, he obtained a Master of Science in Engineering and Computer Science from the same institution in 2023, and is currently pursuing an PhD degree in Systems Engineering and Computer Science in the same university. His research interests include graph neural networks, machine learning and network science.
\endbio

\bio
{}\textbf{Daniel Ratton Figueiredo}
received a BS cum laude degree in Computer Science and MSc degree in Computer and Systems Engineering from the Federal University of Rio de Janeiro (UFRJ) Brazil, in 1996 and 1999, respectively. He received a MSc and PhD degrees in Computer Science from the University of Massachusetts Amherst (UMass) in 2005. He worked as a post-doc researcher at the Swiss Federal Institute of Technology, Lausanne (EPFL). In 2007, he joined the Department of Computer and Systems Engineering (PESC/COPPE) at the Federal University of Rio de Janeiro (UFRJ), Brazil where he currently holds an Associate Professor position. He has a Research Productivity Fellowship granted by CNPq (since 2009) and was member of the Young Scientists Program granted by FAPERJ (from 2010 to 2018). He was a Visiting Fulbright Scholar at Columbia University (6 months in 2017). His current main interests are in Network Science and Graph Learning.
\endbio

\end{document}